\documentclass[a4paper, 10pt, conference]{ieeeconf}      
\usepackage{FG2020}
\usepackage{graphicx}
\usepackage{amsmath}
\usepackage{amsfonts}
\usepackage{booktabs}
\usepackage{makecell}
\usepackage{balance}
\usepackage[nolist]{acronym}

\FGfinalcopy 

\IEEEoverridecommandlockouts                              
\def\FGPaperID{****} 

\title{\LARGE \bf
The FaceChannel: A Light-weight Deep Neural Network for Facial Expression Recognition
}

\author{\parbox{16cm}{\centering
    {\large Pablo~Barros$^1$, Nikhil~Churamani$^2$ and Alessandra Sciutti$^1$}\\
    {\normalsize
    $^1$ Cognitive Architecture for Collaborative Technologies Unit, Istituto Italiano di Tecnologia, Genova, Italy \protect \\
    Email: \{pablo.alvesdebarros, alessandra.sciutti\}@iit.it\\
    $^2$ Department of Computer Science and Technology, University of Cambridge, United Kingdom
    \protect \\
    Email: nikhil.churamani@cl.cam.ac.uk
    }}
}

\begin{document}

\ifFGfinal
\thispagestyle{empty}
\pagestyle{empty}
\else
\author{Anonymous FG2020 submission\\ Paper ID \FGPaperID \\}
\pagestyle{plain}
\fi
\maketitle

\begin{abstract}

Current state-of-the-art models for automatic \acf{FER} are based on very deep neural networks that are difficult to train. This makes it challenging to adapt these models to changing conditions, a requirement from \ac{FER} models given the subjective nature of affect perception and understanding. In this paper, we address this problem by formalising the FaceChannel, a light-weight neural network that has much fewer parameters than common deep neural networks. We perform a series of experiments on different benchmark datasets to demonstrate how the FaceChannel achieves a comparable, if not better, performance, as compared to the current state-of-the-art in \ac{FER}.

\end{abstract}

\section{INTRODUCTION}

It is well-accepted that `basic' emotions are perceived, recognized and commonly understood by people across cultures, around the world~\cite{Ekman1971}. Translating such an idea to universal automatic \acf{FER}, however, is a major challenge. A common obstacle in developing such a solution is that each person might express basic affect differently. Most natural expressions are composed of a series of transitions between several basic affective reactions, occurring in a very short period of time~\cite{cavallo2018emotion}. 
Contributing to the complexity of analyzing facial expressions automatically is the subjective nature of affective expression. Depending on the context of the interaction, the level of engagement and the affective bond between the interaction partners the understanding of an affective exchange between them can change drastically~\cite{Hamann2004,Hess2016}.

One way to address this problem is by formalizing affect in a manner that bounds the categorization ability of a computational system. This requires to choose a highly effective formalization for the task at hand~\cite{griffiths2003iii,barrett2006solving,Afzal2009}. Although there exists extensive research on defining, identifying and understanding human affect, most of the computational models that deal with affect recognition from facial expressions are limited to very limited categorizations. The most commonly used approach is categorization into emotional categories (with universal emotions being the most popular) or using a dimensional representation (usually, arousal and valence), presumably due to the availability of training data that provides such encapsulations.

Most of the current, and effective, solutions for automatic affect recognition are based on extreme generalization, usually employing end-to-end deep learning techniques~\cite{mehta2018facial}. Such models usually learn how to represent affective features from a large number of data samples, using strongly supervised methods~\cite{hazarika2018self,huang2019speech, Kret2013,kollias2019deep, kollias2020analysing}. As a result, these models can extract facial features from a collection of different individuals, which contributes to their generalization of expression representations enabling a universal and automatic \ac{FER} machine. The development of such models was supported by the collection of several ``in-the-wild'' datasets~\cite{dhall2012collecting,mollahosseini2017affectnet,zadeh2018multimodal, zafeiriou2017aff} that provided large amounts of well-labelled data. These datasets usually contain emotion expressions from various multimedia sources ranging from single frames to a few seconds of video material. Because of the availability of large amounts of training data, the performance of deep learning-based solutions forms the state-of-the-art in \ac{FER}, benchmarked on these datasets~\cite{choi2018convolutional,marinoiu20183d,du2019spatio,yang2018deep}.

Most of these models, however, employ large and deep neural networks that demand a lot of computational resources for training. As a result, these models specialize on recognizing emotion expressions under conditions represented in the datasets they are trained with. Thus, when these models are applied to data under different conditions, not represented in the training data, they tend to perform poorly. Retraining these networks to adapt to these novel scenarios is usually helpful in improving the performance of these models. Yet, owing to the large and deep architecture of these models, retraining the entire network with changing conditions is rather expensive. 

Furthermore, once trained, these deep neural models are relatively fast to recognize facial expressions when provided with rich computational resources. With reduced processing power, however, such as in robotic platforms, these models usually are extremely slow and do not support real-time application.

In this paper, we address the problem of using such deep neural networks by formalizing the FaceChannel neural network. This model is an upgraded version of the Multi-Channel Convolution Neural Network, proposed in our previous work~\cite{barros2016developing}.
The FaceChannel is a light-weight convolution neural network, trained from scratch, that presents state-of-the-art results in \ac{FER}. To evaluate our model, we perform a series of experiments with different facial expressions datasets and compare them with the current state-of-art. 

\section{The FaceChannel}

\begin{figure*} 
	\center{\includegraphics[width=1\linewidth]{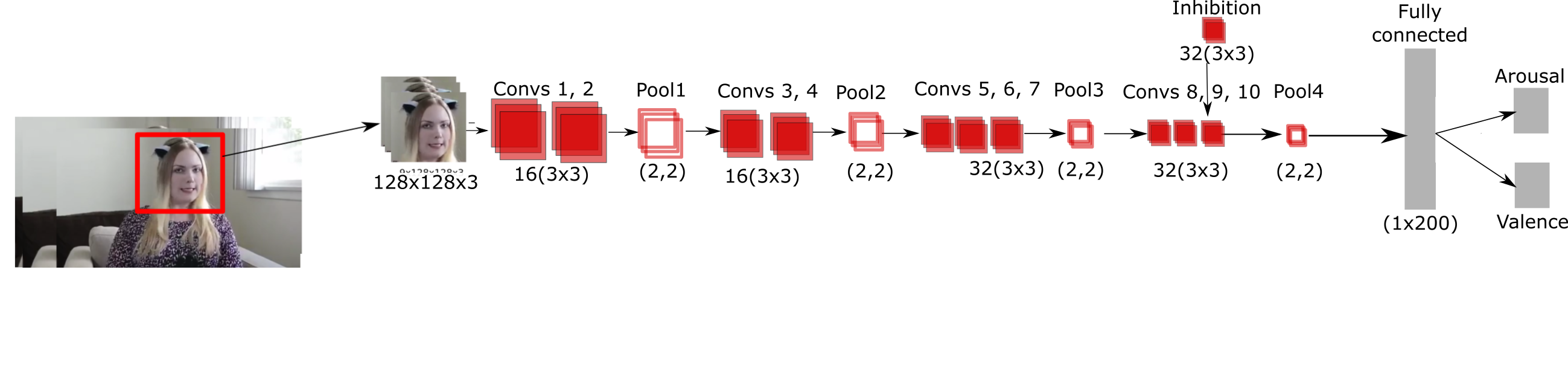}}
	\caption{Detailed architecture and parameters of The FaceChannel.}
	\label{fig:faceChannel}
\end{figure*}


In this paper, we present an updated version of the original FaceChannel proposed in our earlier works~\cite{barros2016developing} that employs a VGG16 model~\cite{VGG2016}-based topology, but with much fewer parameters (see Fig.~\ref{fig:faceChannel}), to improve the robustness of the model. The FaceChannel network has $10$ convolutional layers, including 4 pooling layers. We use batch normalization within each convolutional layer and a dropout function after each pooling layer. Following the original FaceChannel architecture, we apply shunting inhibitory fields~\cite{Fregnac2003} in our last layer. Each shunting neuron $S_{nc}^{xy}$ at the position ($x$,$y$) of the $n^{th}$ receptive field in the $c^{th}$ layer is activated as:

\begin{equation}
S_{nc}^{xy} = \frac{u_{nc}^{xy}}{a_{nc} + I_{nc}^{xy}}
\end{equation}

\noindent where $u_{nc}^{xy}$ is the activation of the common unit in the same position and $I_{nc}^{xy}$ is the activation of the inhibitory neuron. A learned passive decay term, $a_{nc}$ is the same for each shunting inhibitory field. Each convolutional and inhibitory layer of the FaceChannel implements a \textit{ReLu} activation function.

The output of the convolutional layers is fed to a fully connected layer with $200$ units, each one implementing a \textit{ReLu} activation function, which is then fed to an output layer. Our model is trained using a \textit{categorical cross-entropy loss} function. 

As typical for most deep learning models, our FaceChannel has several hyper-parameters that need to be tuned. We optimized our model to maximize the recognition accuracy using a \ac{TPE}~\cite{bergstra2011algorithms} and use the optimal training parameters throughout all of our experiments. The entire network has around $2$ \textit{million} adaptable parameters, which makes it very light-weight as compared to commonly used VGG16-based networks.

\section{Experimental Setup}

To evaluate the FaceChannel, we perform several benchmarking experiments on different \ac{FER} datasets. As some of these datasets do not contain enough data to train a deep neural network, we pre-train our model on the AffectNet dataset~\cite{mollahosseini2017affectnet}. We then \textit{fine-tune} the model with the train split of each dataset and evaluate it following individual evaluation protocols.

\subsection{Datasets}

\begin{figure} 
	\center{\includegraphics[width=1\linewidth]{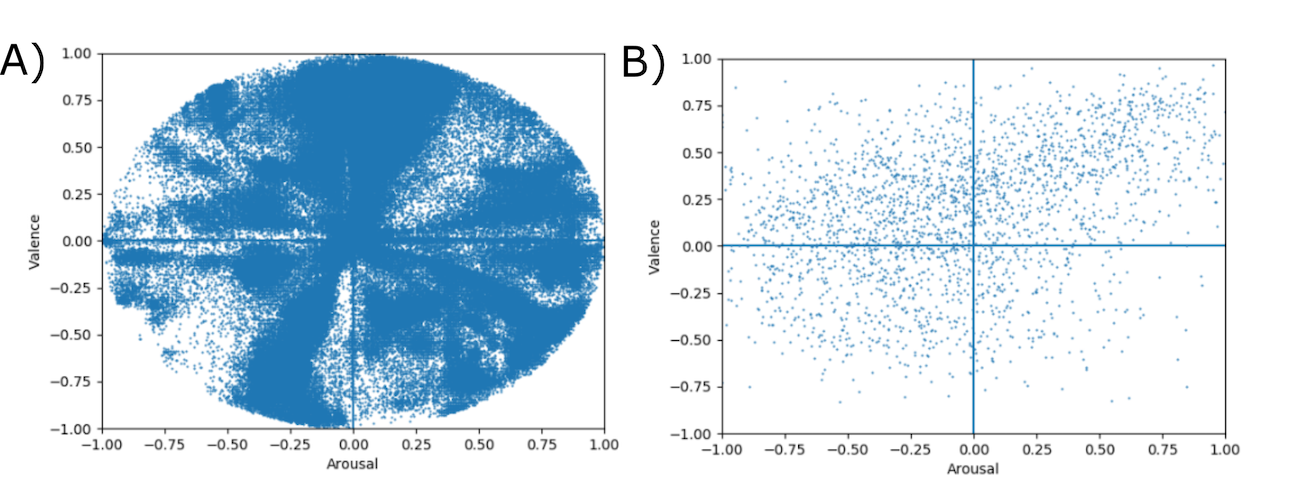}}
	\caption{Annotation distributions for: A) the AffectNet dataset~\cite{mollahosseini2017affectnet} has a high variance on arousal and valence with a large number of data points and B) The continuous expressions of the OMG-Emotion~\cite{barros2018omg} videos cover high arousal and valence spread.}
	\label{fig:dataDistribution}
	\end{figure}

\subsubsection{AffectNet}	
AffectNet~\cite{mollahosseini2017affectnet} consists of more than 400 thousand ``in-the-wild'' images that are manually annotated with seven categorical labels and continuous arousal and valence. As the test-set labels are not publicly available, all our experiments are performed using only the training and validation samples.

As AffcetNet, by far, consists of the best data distribution we experimented with (see Fig.~\ref{fig:dataDistribution} A), we use it to pre-train the FaceChannel for all the other experiments. This also guarantees that the FaceChannel first learns how to extract facial features relevant for \ac{FER}. Fine-tuning the model with each specific dataset allows it to adapt to each of the specific conditions represented in these datasets.

\subsubsection{OMG-Emotion}
We also provide experimentation on the \acf{OMG-Emotion} dataset~\cite{barros2018omg}, which is composed of one minute long YouTube videos annotated taking into consideration continuous emotional behavior. This dataset helps us to evaluate how the model performs on recognising expressions from a particular individual. The videos were selected from YouTube using a crawler technique that uses specific keywords based on long-term emotional scenes such as ``monologues'', ``auditions'', ``dialogues'' and ``emotional scenes'', which guarantees that each video has only one person performing an emotional display. A total of $675$ videos were collected, which sums up to around $10$ hours of audio-visual data. Each utterance on the videos is annotated with two continuous labels, representing arousal and valence. The emotion expressions displayed in the \ac{OMG-Emotion} dataset are heavily impacted by person-specific characteristics that are highlighted by the gradual change of emotional behavior over the entire video. The videos in this dataset cover a diverse range of arousal and valence values as seen in Fig.~\ref{fig:dataDistribution} B.


\subsubsection{FER$+$}
We then evaluate the FaceChannel on the the FER$+$ dataset~\cite{BarsoumICMI2016}. This dataset contains face images, crawled from the internet, with a total of $28,709$ images in the training-set, and $3,589$ in the test-set. This dataset used a crowdsourcing labeling technique, and each image was labeled by at least $10$ different annotators. The obtained labelling distribution is thus provided to train the model. Each label relates to one of the six \textit{universal emotions} (Angry, Disgust, Fear, Happy, Sad, Surprise) and Neutral.


\subsubsection{FABO}
Further, to evaluate the model for controlled environment settings, we train and evaluat it on the Bi-modal Face and Body benchmark dataset FABO~\cite{Gunes2006}. This dataset is composed of recordings of the face and body motion of different subjects using two cameras, one of them capturing only the face and the second one capturing the upper body. Each video contains one subject executing the same expression in a cycle of two to four expressions per video. 

\begin{table}
\caption{Number of videos available for each emotional state in the FABO dataset. Each video has $2$ to $4$ executions of the same expression.}
    \center\begin{tabular}{ c c | c c}
        \toprule
        Emotional State & Videos & Emotional State & Videos      \\ \midrule
        Anger           & $60$  & Happiness       & $28$  \\ 
        Anxiety         & $21$  & Puzzlement      & $46$  \\ 
        Boredom         & $28$  & Sadness         & $16$  \\ 
        Disgust         & $27$  & Surprise        & $13$  \\ 
        Fear            & $22$  & Uncertainty     & $23$  \\\bottomrule
    \end{tabular} 
\label{tab_numberOfFrames}
\end{table}
 
The FABO dataset has annotations for the temporal phase of each video sequence. To create the annotation, six observers label each video independently and then a voting process is executed. We use only upper-body videos that have a voting majority regarding the temporal phase classification, similar to~\cite{Gunes2006}. A total number of $281$ videos are used and are shown in Table \ref{tab_numberOfFrames}. The database contains ten emotional states: ``Anger", ``Anxiety", ``Boredom", ``Disgust", ``Fear", ``Happiness", ``Surprise", ``Puzzlement", ``Sadness" and ``Uncertainty". As necessary for the temporal labeling, only the \textit{apex} state of each sequence is used for training the model. The other frames, present in the remaining temporal phases are grouped into one new category named  ``Neutral" leading to a total of $11$ emotional states to be classified. 


\noindent For all the evaluated datasets, we follow the training and test separation protocols as described by the respective dataset authors to maintain comparability with other proposed models. We pre-process the individual frames to feed them to the FaceChannel. For each video frame, we detect the face of the subject using the Dlib~\footnote{https://pypi.org/project/dlib/} python library. Each face image is then resized to a dimension of $128$x$128$.

\subsection{Metrics}

To measure the performance of the FaceChannel on the respective datasets, we use two metrics: \textit{accuracy},  to recognize categorical emotion expressions, and the \acf{CCC}~\cite{Lawrence1989} between the outputs of the models and the true label to recognize arousal and valence. The \ac{CCC} is computed as:

\begin{equation}
CCC = \frac{2 \rho \sigma_x \sigma_y}{\sigma_{x}^2 + \sigma_{y}^2 + (\mu_x - \mu_y)^2}
\label{eq:ccc}
\end{equation}

\noindent where $\rho$ is the Pearson's Correlation Coefficient between model prediction labels and the annotations, $\mu_x$ and $\mu_y$ denote the mean for model predictions and the annotations and $\sigma_{x}^2$ and $\sigma_{y}^2$ are the corresponding variances. The \ac{CCC} metric allows us to have a direct comparison with the annotations available on the \ac{OMG-Emotion} dataset. The use of \ac{CCC} as the main objective measurement allows us to take into consideration the subjectivity of the perceived emotions for each annotator when evaluating the performance of our models.

Both metrics are in accordance with the experimental protocols defined by each of the individual dataset authors.

\section{Results}

\begin{table}[b]
    \caption{\acf{CCC}, for arousal and valence, and the Categorical Accuracy when evaluating the FaceChannel with the AffectNet, \ac{OMG-Emotion}, FER$+$, and FABO datasets.}
    \center 
    \setlength\tabcolsep{4pt}
    \footnotesize{
    \begin{tabular}{ c |  c | c | c | c }\toprule

    \textbf{Model}                          & \textbf{Arousal}  & \textbf{Valence} & \textbf{Model} & \textbf{Accuracy}  \\\midrule
        
        \multicolumn{3}{c|}{\textbf{AffectNet}~\cite{mollahosseini2017affectnet}}  & \multicolumn{2}{c}{\textbf{FER$+$}~\cite{BarsoumICMI2016}}\\ \midrule
        
        AlexNet~\cite{krizhevsky2012imagenet}   & 0.34              & 0.60  & CNN VGG13~\cite{BarsoumICMI2016}  & 84.98\%  \\
        MobileNet~\cite{hewitt2018cnn} & 0.48 & 0.57 & SHCNN~\cite{miao2019recognizing} & 86.54 \\
        VGGFace~\cite{lindt2019facial}  & 0.40 & 0.48 & TFE-JL~\cite{li2018facial}& 84.3\\
        \textbf{VGGFace+GAN~\cite{kollias2020deep}}  & \textbf{0.54} & \textbf{0.62} & ESR-9~\cite{siqueira2020efficient}&87.15 \\
        
        Face Channel                   & 0.46     & 0.61  & \textbf{Face Channel} & \textbf{90.50\%}  \\\midrule
        
        \multicolumn{3}{c|}{\textbf{OMG-Emotion}~\cite{barros2018omg}} & \multicolumn{2}{c}{\textbf{FABO}~\cite{Gunes2006}}\\\midrule
        
        \textbf{Zheng~et~al.~\cite{zheng2018multimodal}} & \textbf{0.35} & \textbf{0.49} & \makecell{Temporal\\Normalization~\cite{Chen2013}}           &  66.50\%     \\ 
        
        Huang~et~al.~\cite{huang2019speech}     & 0.31              & 0.45  & Bag~of~Words~\cite{Chen2013}           & 59.00\%   \\
        Peng~et~al.~\cite{peng2018deep}         & 0.24              & 0.43  & SVM~\cite{Gunes2009}~           & 32.49\%   \\    
        
        Deng~et~al.~\cite{deng2018multimodal}   & 0.27              & 0.35  & Adaboost~\cite{Gunes2009}           & 35.22\%   \\ 
        FaceChannel                             &  0.32             & 0.46  & \textbf{Face channel} & \textbf{80.54\%} \\\bottomrule
    \end{tabular} 
\label{tab:AllExperiments}
}
\end{table}

Although, the AffectNet corpus is very popular, not many researchers report the performance of arousal and valence prediction on its validation set. This is probably the case as most of the research uses the AffectNet dataset to pre-train neural models for generalization tasks in other datasets, without reporting the performance on the AffectNet itself.  The baseline provided by the authors uses an AlexNet-based Convolutional Neural Network~\cite{krizhevsky2012imagenet} re-trained to recognize arousal and valence. A similar approach is reported by Hewitt and Gunes~\cite{hewitt2018cnn}, but using a much reduced neural network, to be deployed in a smart-device. Lindt et al. \cite{lindt2019facial} reports experiments using the VGGFace, a variant of the VGG16 network pre-trained for face identification. Kollias et al. \cite{kollias2020deep} proposed a novel training mechanics, where it augmented the trainint set of the AffectNet using a Generative Adversarial Network (GAN), and obtained the best reported accuracy on this corpus, achieving $0.54$ CCC for arousal and $0.62$ CCC for valence.  Our FaceChannel provides an improved performance when compared to most of these results, reported in Table~\ref{tab:AllExperiments}, achieving a \ac{CCC} of $0.46$ for arousal and $0.61$ for valence. Different from the work of  Kollias et al. \cite{kollias2020deep}, we train our model using only the available training set portion, and expect these results to improve when training on an augmented training set.


 The performance of the FaceChannel, reported in Table~\ref{tab:AllExperiments}, is very similar when compared to the current state-of-the-art results on the \ac{OMG-Emotion} dataset, as reported by the winners of the \ac{OMG-Emotion} challenge where the dataset was proposed~\cite{zheng2018multimodal,peng2018deep,deng2018multimodal}. All these models also reported the use of pre-training of uni-sensory convolutional channels to achieve such results, but employed deep networks with much more parameters to be fine-tuned in an end-to-end manner. The use of attention mechanisms~\cite{zheng2018multimodal} to process the continuous expressions in the videos presented the best results of the challenge, achieving a \ac{CCC} of $0.35$ for arousal and $0.49$ for valence. Temporal pooling, implemented as bi-directional \acfp{LSTM} , achieved the second place, with a \ac{CCC} of $0.24$ for arousal and $0.43$ for valence. The late-fusion of facial expressions, speech signals, and text information reached the third-best result, with a \ac{CCC} of $0.27$ for arousal and $0.35$ for valence. The complex attention-based network proposed by Huang~et~al.~\cite{huang2019speech} was able to achieve a \ac{CCC} of $0.31$ in arousal and $0.45$ in valence, using only visual information.
 
 When trained and evaluated with the FER+ model, our FaceChannel provides improved results as compared to those reported by the dataset authors~\cite{BarsoumICMI2016}. They employ a deep neural network based on the VGG13 model, trained using different label-averaging schemes. Their best results are achieved using the labels as a probability distribution, which is the same strategy we used. We outperform their result by almost $6\%$ as reported in Table \ref{tab:AllExperiments}. We also outperform the results reported in Miao et al.~\cite{miao2019recognizing}, Li et al.~\cite{li2018facial}, and Siqueira et al.~\cite{siqueira2020efficient} which emply different type of complex neural networks to learn facial expressions.
 
 Our model achieves higher accuracy for the experiments with the FABO dataset as well when compared with the state-of-the-art for the dataset~\cite{Chen2013}. They report an approach based on recognizing each video-frame, similar to ours. The results reported by Gunes~et~al.~\cite{Gunes2009} for Adaboost and SVM-based implementations are reported using a frame-based accuracy. Our FaceChannel outperforms both models, as illustrated in Table \ref{tab:AllExperiments}.

\section{Conclusions}

In this paper, we present a formalization of the FaceChannel neural network for \acf{FER}. The network, an optimized version of the VGG16, has much fewer parameters, which reduces the training and fine-tuning efforts. That makes our model ideal for adapting to specific characteristics of affect perception, which may differ based on where the network is applied.

We perform a series of experiments to demonstrate the ability of the network to adapt to different scenarios, represented here by the different datasets. Each dataset consists of a specific constraint when representing affect from facial expressions. Our model demonstrated to be at par, if not better, than the state-of-the-art solutions reported on these datasets. To guarantee the reproducibility and dissemination of our model, we have made it fully available on GitHub\footnote{https://github.com/pablovin/AffectiveMemoryFramework}.

In the future, we plan to study the application of our model on platforms with reduced data processing capabilities, such as on robots. Also, we plan to optimize the model further, by investigating different transfer learning methods, to make it even more robust to re-adaptation.


\balance

\bibliographystyle{ieeetr}
\bibliography{bib}

\begin{acronym}
\acro{CCC}{Concordance Correlation Coefficient}
\acro{FER}{Facial Expression Recognition}
\acro{LSTM}{Long Short-Term Memory}
\acro{OMG-Emotion}{One Minute Gradual Emotion Recognition}
\acro{TPE}{Tree-structured Parzen Estimator}
\end{acronym}
\end{document}